\documentclass{article}
\usepackage{spconf,amsmath,graphicx}

\usepackage{enumitem}
\usepackage{multirow}
\usepackage{amsfonts}
\setlist{nosep, leftmargin=14pt}

\usepackage{mwe} 

\usepackage{hyperref}
\hypersetup{
	colorlinks,
	linkcolor=[rgb]{1.0, 0.0, 0.0},
	citecolor=[rgb]{1.0, 0.2, 0.2},
	urlcolor =[rgb]{0.0, 0.0, 0.8}
}

\newcommand{\featureaugmenter}{\textsc{EmbAugmenter}}
\newcommand{\ind}{\perp\!\!\!\!\perp}

\title{Embedding Space Augmentation for Weakly Supervised Learning in Whole-Slide Images}

\name{Imaad Zaffar$^{1,\star}$ \qquad Guillaume Jaume$^{2,3,4,5,\star}$ \qquad Nasir Rajpoot$^{6,7,8}$ \qquad Faisal Mahmood$^{2,3,4,5}$}

\address{
$^{1}$ Department of Computer Science, University College London, UK \\
$^{2}$Department of Pathology, Brigham and Women's Hospital, Harvard Medical School, Boston, MA, USA \\
$^{3}$Department of Pathology, Massachusetts General Hospital, Harvard Medical School, Boston, MA, USA \\
$^{4}$Cancer Program, Broad Institute of Harvard and MIT, Cambridge, MA, USA \\
$^{5}$Data Science Program, Dana-Farber Cancer Institute, Boston, MA, USA \\
$^{6}$ Tissue Image Analytics Centre, Department of Computer Science, University of Warwick, Coventry, UK \\
$^{7}$ Department of Pathology, University Hospitals Coventry and Warwickshire NHS Trust, Coventry, UK \\
$^{8}$ The Alan Turing Institute, London, UK\\
$\star$ denotes equal contribution
}

\begin{document}

\maketitle
\begin{abstract}
Multiple Instance Learning (MIL) is a widely employed framework for learning on gigapixel whole-slide images (WSIs) from WSI-level annotations. In most MIL based analytical pipelines for WSI-level analysis, the WSIs are often divided into patches and deep features for patches ({\em i.e.}, patch embeddings) are extracted \emph{prior} to training to reduce the overall computational cost and cope with the GPUs' limited RAM.
To overcome this limitation, we present $\featureaugmenter$, a data augmentation generative adversarial network (DA-GAN) that can synthesize data augmentations in the \emph{embedding space} rather than in the \emph{pixel space}, thereby significantly reducing the computational requirements. Experiments on the SICAPv2 dataset show that our approach outperforms MIL without augmentation and is on par with traditional patch-level augmentation for MIL training while being substantially faster. 
\end{abstract}

\section{Introduction}
\label{sec:intro}

Computational pathology has made significant progress in recent years with new methods capable of classifying high-dimensional whole-slide images (WSI) of the order of 100,000 $\times$ 100,000 pixels~\cite{lu2021ai,bilal2021development}. Most of these successes are based on weakly supervised learning, particularly Multiple Instance Learning (MIL). In MIL, each WSI is associated with a label that needs to be inferred from the set of patches that form the WSI in a two-step process. First, each patch is pre-processed by a pre-trained feature extractor to reduce the dimensionality of the patches (\emph{e.g.,} typically from a 256$\times$256 patch to 1024 dimensional feature vector that makes the patch embedding). The feature extractor can either be pre-trained on an auxiliary task~\cite{lu2021data} or based on self-supervised learning~\cite{koohbanani2021self,chen2022scaling}. Due to the large number of patches per WSI (can be $> 10,000$), this step is computationally intensive and is typically only performed once. Then, in the second step, a neural network combines the low-dimensional patch embeddings into a slide-level representation used for classification. 

\begin{figure}[t]
\centering
\includegraphics[width=0.49\textwidth]{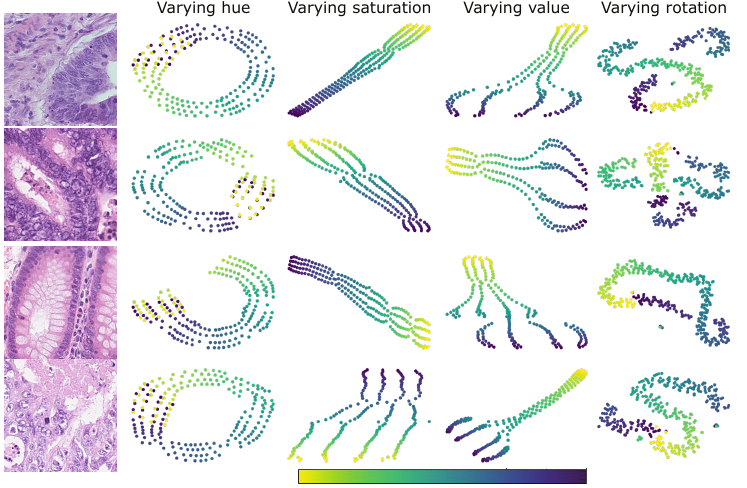}
\caption{Low-dimensional visualization of the ResNet50 embedding space when varying image hue, saturation, value, and rotation for four example patch images.} 
\label{fig:teaser}
\end{figure}	

\begin{figure*}[t]
\centering
\includegraphics[width=0.95\textwidth]{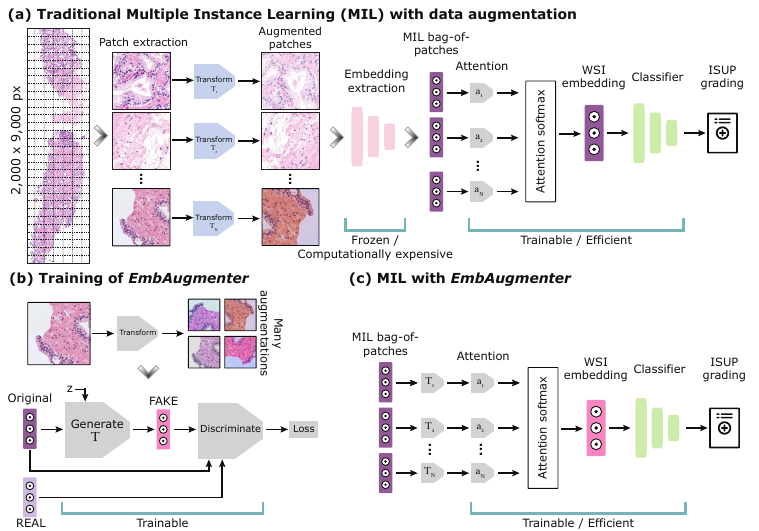}
\caption{\textbf{MIL training using traditional data augmentation {\em vs} the proposed method:} ({\em a}) Traditional data augmentation operates in pixel space at a high computational cost; ({\em b}) Instead, we train a Data-Augmentation GAN to synthesize patch embeddings; ({\em c}) Arbitrary patch embeddings can be generated during MIL training to enhance data variability.} 
\label{fig:overview}
\end{figure*}	

Although MIL methods are widely used, pre-extracting patch embeddings beforehand is computationally demanding and time-consuming for incorporating data augmentation -- a technique often used in computer vision to help reduce overfitting and increase generalization. In MIL, patch embeddings would have to be extracted as many times as a new augmentation is needed. Since patch embedding extraction is the most time-consuming part of the training, the total cost quickly becomes very expensive~\cite{tellez19}. In addition, in this training setting, only a finite number of embeddings can be extracted, which limits the augmentation variability. 

Instead, we propose to generate augmentations in the \emph{embedding space} by synthesizing variations of the patch embeddings. In this way, we only need to extract patch embeddings once on the original images and augment them during training. Specifically, we employ a data augmentation generative adversarial network (DA-GAN) to learn the distribution of patch embeddings (see Figure~\ref{fig:teaser}). After training, the GAN generator can be re-instantiated to generate entirely new augmentations from the original embeddings.

Specifically, our contributions are: (1) We propose a novel GAN-based $\featureaugmenter$ that learns to synthesize rich augmentations of \emph{patch embeddings} at a minimal cost; (2) We show that MIL training with the $\featureaugmenter$ outperforms MIL training without augmentation at a fraction of the computational cost of original augmentations and is comparable to traditional patch-level augmentation on the SICAPv2 dataset for ISUP grading of prostate biopsies.

\section{The Proposed Method}

In this section, we present our approach for enabling embedding space augmentation during MIL training. An overview of the method is shown in Fig.~\ref{fig:overview}.

\subsection{Patch embedding augmentation}

\begin{figure}[t]
\centering
\includegraphics[width=0.48\textwidth]{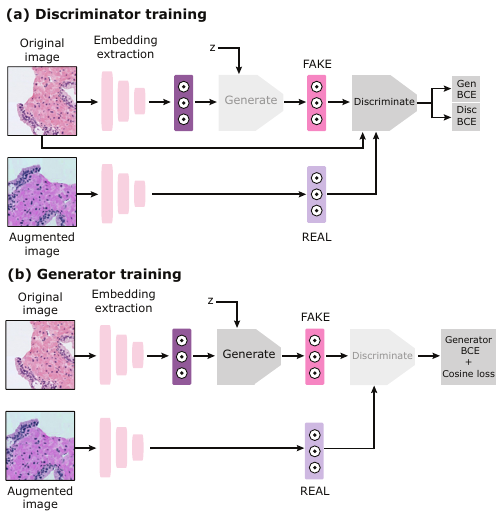}
\caption{\textbf{Generative embedding space augmentation}. (a) The discriminator is trained to distinguish real from synthetic patch embeddings. (b) The generator is trained to synthesize plausible patch augmentations.} 
\label{fig:gan_training}
\end{figure}

We first decompose an input WSI $X$ into a bag of $N$ patches, denoted as $X = \{\mathbf{x}_1, ..., \mathbf{x}_N\}$. We then employ a feature extractor $f(\cdot)$ to map each patch $\mathbf{x}_i$ into a patch embedding as $\mathbf{h}_i = f(\mathbf{x}_i) \in \mathbb{R}^d$. Our goal is then to synthesize embeddings of patch augmentations. For this purpose, we introduce $\featureaugmenter$, a data augmentation GAN that consists of a generator denoted as $\mathcal{T}$ and of a discriminator denoted as $D$. Given a patch embedding $\mathbf{h}_i = f(\mathbf{x}_i)$ and a randomly sampled latent vector $\mathbf{z} \in \mathbb{R}^d$, we learn a network $\mathcal{T}: (\mathbf{z}, \mathbf{h}_i) \mapsto \mathbf{\tilde{h}}_i$ that can synthesize patch embeddings resembling the true embeddings of patch augmentations. Similar to the traditional GAN training~\cite{goodfellow14}, the generator and discriminator play a min-max game, where the discriminator tries to identify real from fake samples while the generator iteratively learns better and closer embeddings (see Fig.~\ref{fig:gan_training}).

We propose two variants of the GAN generator and discriminator with different model expressivity. First, a model that assumes that the embedding factors are independent of each other, \emph{i.e.,} $\mathbf{\tilde{h}}_i^{(j)} \ind \mathbf{\tilde{h}}_i^{(j+1)},\; j \in \{1:d\}$. The second variant models all-to-all interactions between the patch embedding factors and $z$. Formally, the generator $\mathcal{T}$ is expressed as,
\begin{align}
    \mathcal{T}_{\text{Exp}}(\mathbf{z}, \mathbf{h}_i) &= \text{MLP}_{\text{Exp}}(\mathbf{z} \parallel \mathbf{h}_i) \\
    \mathcal{T}_{\text{Ind}}(\mathbf{z}, \mathbf{h}_i) &= \Big|\Big |_{j=1}^d\Big( \text{MLP}_{\text{Ind}}(\mathbf{z}^{(j)} \parallel \mathbf{h}_i^{(j)})\Big) 
\end{align}
where $||$ denotes the concatenation operation, MLP denotes a multi-layer perceptron, $\text{MLP}_{\text{Ind}}$ denotes the independent model (Ind), and $\text{MLP}_{\text{Exp}}$ denotes the expressive variant (Exp). The discriminator is defined analogously. The generator loss is composed of two terms: the cosine similarity between the true and fake patch embeddings and the discriminator binary cross-entropy (BCE). As in regular GAN training, the discriminator loss is simply a BCE term. Prior to GAN training, true patch embeddings are extracted using  patch augmentations based on random rotation, color jittering, and zoom in/out. In essence, $\featureaugmenter$ is similar to a Pix2Pix model~\cite{isola17} where the pixels would be replaced by patch embedding factors. 

\subsection{Embedding space augmented MIL training}

We now present how $\featureaugmenter$ can be integrated into MIL training. Each WSI $X$ is associated with a label $y$ that we aim to predict. In this work, we employ an attention-based MIL model~\cite{ilse18attention}. Specifically, after patch-level feature extraction, we increase the data variability by further augmenting the patch embeddings $\mathbf{h}_{i=1{\dots}N}$ using our proposed $\featureaugmenter$ yielding to $\mathbf{\Tilde{h}}_{i=1{\dots}N} = \{\mathcal{T}(\mathbf{h})\}_{i=1{\dots}N}$, where $\mathcal{T}(\cdot)$ is the GAN generator. In a second step, a neural network, denoted as $g(\cdot)$, combines the embeddings into a WSI-level embedding $h_{\text{WSI}} \in \mathbb{R}^{d_{\text{WSI}}}$, that is finally fed to a predictor, denoted as $c(\cdot)$, for classifying the WSI. These steps can be summarized as,
\begin{equation} \label{eq:mil_funcs}
    \hat{\mathbf{y}} = c \bigg( g \Big( \{ \mathcal{T}\big(f(\mathbf{x}_1)\big), \dots , \mathcal{T}\big(f(\mathbf{x}_N)\big) \} \Big) \bigg)
\end{equation}
where $\hat{\mathbf{y}}$ denotes the WSI prediction. The augmented patch embeddings are combined into a WSI representation using an attention mechanism as $g(\cdot) = \sum_{i=1}^N a_i \mathbf{h}_i$, where $\{a_i\}_{i={1:N}}$ are gated-attention weights~\cite{bahdanau2014neural,ilse18attention}.  Using this approach, the time-consuming step of extracting features on pixel-augmented patches is replaced by an efficient \emph{embedding} space augmentation. An overview of this process is depicted in Fig.~\ref{fig:overview}(c).

\section{Experimental Results}

\newcommand{\std}[1]{{\scriptsize $\pm$#1}}
\begin{table*}[t]
    \centering
    \caption{Performance of the $\featureaugmenter$ tested on the SICAP dataset.} 

    \begin{tabular}{l|cccc}
 
     Augmentation & Accuracy ($\%$, $\uparrow$) & $\kappa^2$ ($\%$, $\uparrow$) & NLL ($\downarrow$) \\
     \hline
     No augmentation  & $46.5$ \std{22.4} & $0.55$ \std{0.39} & $2.97$ \std{1.45} \\

     Patch augmentation & $\mathbf{52.3}$ \std{10.8} & $\mathbf{0.76}$ \std{0.07} & $\mathbf{1.38}$ \std{0.40} \\

    $\text{MLP}_{\text{Ind.}}$, $\featureaugmenter$ & $\mathbf{52.3}$ \std{9.0}  & ${0.71}$ \std{0.11} & $1.62$ \std{0.58} \\

    $\text{MLP}_{\text{Exp.}}$, $\featureaugmenter$ & $\underline{48.4}$ \std{12.9} &  $\underline{0.74}$ \std{0.11} & $\underline{1.50}$ \std{0.48} \\
    \end{tabular}
    \label{tab:results}
\end{table*}

We benchmark $\featureaugmenter$ on the SICAPv2 dataset for ISUP grading of prostate biopsies (5-class problem). SICAPv2 comprises 155 WSIs of varying shapes and sizes processed at $10\times$ magnification with an average of 129 256$\times$256 patches per WSI. 5-fold cross-validation was employed with a 60\%, 20\%, 20\% train, validation and test split. A different $\featureaugmenter$ was trained for each fold to avoid transductive data leakage between training and testing. Each $\featureaugmenter$ training used 48,552 $(\mathbf{h}_i, \mathcal{T}\mathbf({h}_i))$ pairs. We benchmark $\featureaugmenter$ in terms of classification performance -- measured with the accuracy, quadratic kappa score ($\kappa^2$), and negative log-likelihood (NLL) -- and computational time.

We compare $\featureaugmenter$ against two baselines. First, a baseline where the MIL model is trained without augmentation (referred to as \emph{No augmentation} in Table~\ref{tab:results}), and second a baseline that uses traditional pre-extracted patch-level augmentations (referred to as \emph{Patch augmentation}). In our experiments, we used five augmentations per patch. To ensure a fair comparison, the three approaches are based on the same model hyperparameters and only differ in terms of the method of augmentation. All methods were optimized to determine the optimal learning rate and weight decay. The code was implemented in PyTorch and optimized with Adam.

The $\featureaugmenter$ generator and discriminator are both based on MLPs: $\text{MLP}_{\text{Exp}}$ is a 6-layer encoder/decoder MLP  with 256 bottleneck dimensions and $\text{MLP}_{\text{Ind}}$ is a 2-layer MLP with 4 hidden dimensions. The feature extractor $f(\cdot)$ is using ResNet50~\cite{he2016deep} features pre-trained on ImageNet as previously proposed in~\cite{lu2021data,shao2021transmil}. The attention in $g(\cdot)$ uses a gated mechanism with 2-layer MLPs to map the 1024 patch embedding dimensions to a single attention weight. The classifier $c(\cdot)$ uses a 2-layer MLP with 256 hidden dimensions.

Table~\ref{tab:results} presents classification results on SICAPv2. Including augmentation during training (at both patch and embedding levels) leads to a significant performance boost, \emph{e.g.,} absolute gain of $+5.8\%$ in accuracy with and without patch augmentation. While the embedding space features may appear to be independent to some extent (lower NLL loss than the baseline without augmentation), having a more expressive model that captures interactions between \emph{all} the features further increases performance. Interestingly, $\text{MLP}_{\text{Exp.}}$, $\featureaugmenter$ leads to only slightly lower performance compared to traditional patch-level augmentation both in terms of classification performance and NLL. We hypothesize that, while MIL training with embedding augmentations can be generated as many times as needed at training time, they do not capture the entire spectrum of true patch augmentations.  

The computational gain of embedding space augmentation over pixel space is more than $300\times$ for both $\text{MLP}_{\text{Exp}}$ and $\text{MLP}_{\text{Ind}}$. 

\section{Conclusions and Future Directions}

In this paper, we proposed a new technique for image data augmentation in the embedding space. The proposed $\featureaugmenter$ is particularly valuable for training MIL methods that rely on pre-extracted patch embedding representations. With the $\featureaugmenter$, new augmentations can be generated during training at each epoch, thus increasing the variability of the data in an efficient manner. In the future, this method can be tested on larger datasets with thousands of WSIs. $\featureaugmenter$ can also be conditioned by the augmentation type, enabling application- and training-specific augmentations.  

\bibliographystyle{IEEEbib}
\bibliography{refs}

\section{Acknowledgments}

A large part of this work was conducted while one of the first authors (IZ) was visiting the Harvard Medical School. NR is founder, CSO and Director of Histofy, a UK based company. 


\end{document}